\documentclass[12pt]{article}
\usepackage[english]{babel}
\usepackage{picinpar,graphicx,amsmath,amsthm,amsfonts,amssymb,subcaption,CJKutf8,float,hyperref}

\begin{document} 
	\title{Enhancing Solution Efficiency in Reinforcement Learning: Leveraging Sub-GFlowNet and Entropy Integration}
	\author{Siyi He} 
	\maketitle  
	\section*{Abstract}
	Traditional reinforcement learning often struggles to generate diverse, high-reward solutions, especially in domains like drug design and black-box function optimization. Markov Chain Monte Carlo (MCMC) methods provide an alternative method of RL in candidate selection but suffer from high computational costs and limited candidate diversity exploration capabilities. In response, GFlowNet, a novel neural network architecture, was introduced to model complex system dynamics and generate diverse high-reward trajectories. To further enhance this approach, this paper proposes improvements to GFlowNet by introducing a new loss function and refining the training objective associated with sub-GFlowNet. These enhancements aim to integrate entropy and leverage network structure characteristics, improving both candidate diversity and computational efficiency. We demonstrated the superiority of the refined GFlowNet over traditional methods by empirical results from hypergrid experiments and molecule synthesis tasks. The findings underscore the effectiveness of incorporating entropy and exploiting network structure properties in solution generation in molecule synthesis as well as diverse experimental designs.
	
	\section{Introduction}
	In drug design and functional optimization across various fields, it is crucial to generate a diverse array of high-reward candidate solutions. For example, in protein molecule synthesis, this diverse array enables more effective selection of potential candidates, boosting high-score protein generation. Traditional reinforcement learning methods struggle with this challenge. They often focus too much on a single high-reward solution. As a result, they fail to explore other potential high-reward solutions. However, in certain applications, such as drug discovery, it's crucial to sample from multiple high-reward solutions to increase the probability of finding effective drugs. Existing methods such as Markov Chain Monte Carlo (MCMC) can convert energy functions into generating distributions. However, MCMC is often computationally expensive and limited to local exploration. Standard reinforcement learning methods tend to converge to a single highest-reward solution, thus hindering the generation of diverse high-reward candidate solutions. As a substitute of existing methods, GFlowNet is introduced as a method to model the dynamics of complex systems like chemical reactions or social interactions. To overcome current limitations, GFlowNet utilizes a neural network architecture to capture probabilistic relationships among variables over time. By training on observed data, GFlowNet learns to produce trajectories resembling the system's behavior. Hence, GFlowNet is uniquely capable of handling both linear and branching trajectories for more accurate modeling.
	
	Initially presented by Bengio et al. (2021), GFlowNet addresses the challenge of generating diverse objects based on a given reward function. It aims to overcome the high training costs and limited exploration capabilities of traditional methods such as MCMC. GFlowNet conceptualizes the generation process as a flow network. Therefore, GFlowNet is able to transform trajectory sets into flow networks to enhance efficiency and diversity in generated samples. Subsequent enhancements have been proposed to refine its performance: Malkin et al. (2022) introduced trajectory balance to expedite learning and convergence, while Zhang et al. (2022) developed EB-GFN, integrating energy models to effectively learn from energy distributions. Ekbote et al. (2022) adapted EB-GFN for multivariate joint distributions, resulting in JEBGFNs. JEBGFNs significantly enhances efficiency and diversity in generating antimicrobial peptides. Madan et al. (2022) proposed sub-trajectory balance to better leverage local information, aiming to balance bias and variance. Similarly, Shen et al. (2022) introduced guided trajectory balance (GTB) to address local credit assignment issues. Further refinements by Pan et al. (2023) led to FL-GFN, reparameterizing the state flow function to accumulate rewards. FL-GEN successfully surpasses previous methods. In summary, the aforementioned studies all focus on directly improving GFlowNet itself to enhance its efficiency.
	
	For enhanced training effectiveness, scholars have taken different approaches, focusing on preparatory work for using GFlowNet. Approaches including refining evaluation strategies and flow parameterization are employed to improve GFlowNet's sampling efficiency. Shen et al. (2022) introduced the PRT method (Priority Replay Training) to better evaluate GFlowNet. In detail, PRT compares known sample distributions with target reward distributions. Consequently, SSR method is proposed for prioritizing high-reward samples during training. Yet,  Rector-Brooks et al. (2023) addressed the lack of systematic methods for exploring optimal training trajectories by introducing the TS-GFN (Thompson Sampling GFlowNet) algorithm. This strategy enhanced the state space exploration and broadened the range of generated candidates.
	
	While preparatory work for GFlowNet are emphasized, other scholars have focused on enhancing GFlowNet from other perspectives, such as expanding its application scope or examining it through new theoretical frameworks. Lahlou et al. (2023) extended GFlowNet to continuous and mixed spaces. More specifically, the extension adapted components like reward function matching and balance conditions for superior results. Deleu and Bengio (2023) positioned GFlowNet within the MCMC framework, highlighting similarities and differences between GFlowNet and MCMC. This handling of GFlowNet provides a theoretical summary of its capabilities. Bengio et al. (2023) further provided a comprehensive overview, showcasing GFlowNet's abilities in estimating distributions, conditional probabilities, entropy, mutual information, extensions to stochastic environments and modular energy functions.
	
	Applications of GFlowNet span various biological and chemical tasks, including drug discovery, small molecule design and molecular generation. Jain et al. (2022) used GFlowNet to ensure diversity in candidate molecules for drug discovery, while Nica et al. (2022) evaluated its performance in small molecule design tasks. Jain et al. (2023) introduced Multi-Objective GFlowNets (MOGFNs) for optimizing multiple conflicting objectives in molecular generation tasks. In generative modeling, Zhang et al. (2022) explored connections between existing deep generative models and GFlowNets, proposing MLE-GFN to improve generative modeling methods. Subsequently, MLE-GFN demonstrated superior performance in most benchmark distributions. Additionally, in computer science, GFlowNet has been applied to optimizing scheduling operations in computational graphs. Zhang et al. (2023) used GFlowNet to sample from proxy metrics for optimizing schedules, while Jain et al. (2023) applied GFlowNet to modeling, hypothesis generation, and experimental design in experimental science. GFlowNets application extends to causal inference as well. For instance, Li et al. (2022) proposing GFlowCausal for learning DAGs from observational data. Besides, Emezue et al. (2023) and Deleu et al. (2023) introducing JSP-GFN for approximating Bayesian network structures and parameters. The advantages of JSP-GFN are shown in both simulated and real data. Overall, GFlowNet has shown significant potential and versatility across various domains. 
	
	The current research gap in GFlowNet lies in its excessive focus on linear structures within the existing loss functions. Essentially, current approaches treat the GFlowNet loss function as a summation of loss functions from multiple Markov chains. Therefore, present methods unavoidably overlook the characteristics of network structures. Furthermore, the summation of loss functions rarely incorporates weighting. While some scholars consider trajectory length as a criterion for weighting the loss function, this approach still predominantly reflects the characteristics of linear structures. In other words, this method is a typical example about oversight of network structural features. Additionally, existing weighting scheme includes all sub-trajectories in the computation without filtering based on the inclusion of substructures.
	
	This study proposes a method by integrating network structures into the calculation of the loss function. Specifically, the overall GFlowNet loss function is decomposed into secondary sub-GFlowNet loss functions. The entropy of the sub-GFlowNet serves as a weighting criterion of loss funtions. The proposition of this weighting scheme is inspired by the similarity between GFlowNet and decision trees. Additionally, this study only includes points with special branching into the computation of sub-GFlowNet loss functions. Hence, the proposed approach partially addressed the issue of substructure selection. In the hypergrid environment experiments, grids of dimensions 2, 3, and 4 with varying lengths were investigated. The proposed sub-GFlowNet loss function demonstrates accelerated convergence rates and reduced terminal L1 distance between empirical and real distributions. The improvement is particularly evident in 2-dimensional grid experiments. The convergence speed of the sub-GFlowNet loss slightly decreased in higher dimensions. However, as the dimensionality increases, the sub-GFlowNet loss continues to outperform competing loss functions. Moving to molecule synthesis, the investigation aims to generate molecular structures with minimized binding affinity to the sEH protein inhibitor. Through graph representation and junction trees, the study navigates a vast action space with trajectory lengths and molecular complexities, yielding promising results. 
	The sub-trajectory balance GFlowNet showcases decreased diversity in molecule generation. Nevertheless, the proposed GFlowNet notably demonstrates superior convergence and reward attainment.
	
	The structure of this paper is as follows: the preliminaries section will cover the foundational background of GFlowNet. The model section will introduce the enhanced GFlowNet and the novel loss function calculation approach developed in this study. The experiment section will involve numerical simulations in the hypergrid environment and comparative tests on actual data from protein synthesis. Finally, the conclusion section will summarize the entire paper.
	
	Overall, these findings underscore the effectiveness of the sub-GFlowNet loss function in guiding trajectory optimization across both synthetic and real-world scenarios. The proposed approach offers accelerated convergence and enhanced diversity in solution space exploration.
	\section{Preliminaries}       
	\subsection{Data Structure}
	
	Recall GFlowNet aims to capture the essence of sequential decision-making in object construction processes. Under predefined constraints, this framework utilizes a user-specified deterministic Markov Decision Process (MDP) to formulate a generative policy.
	
	The MDP configuration comprises the following components: a state space denoted by $\mathcal{S}$, a set of permissible actions $\mathcal{A}(s)$ corresponding to each state $s$ (The sum of $\mathcal{A}(s)$ is denoted as $\mathbb{A}$), a deterministic transition mechanism represented by $\mathcal{S} \times \mathbb{A} \rightarrow \mathcal{S}$, and a reward function $R$. To illustrate $\mathcal{S} \times \mathbb{A} \rightarrow \mathcal{S}$, when a state $s$ in $\mathcal{S}$ is applied an action $s\rightarrow s'$ in $\mathcal{A}(s)$, $s$ will deterministically move to another state $s'\in \mathcal{S}$. Reward function describes the certain goal characteristic of candidates which researchers pay attention to. For instance, when generating small molecules, the researcher may focus on the binding energy of a molecule to a particular protein target. Consequently, the function representing the binding energy will be reward function in this task.  
	
	Within the GFlowNet framework, the MDP is represented as a structured graph known as a flow network. In this network, nodes represent states, and directed edges are determined by the MDP transition dynamics. Nodes with incoming edges are children. Nodes with outgoing edges are considered as parents, whereas those without are termed terminal states or sinks $s_f$.
	
	A trajectory in the context of GFlowNet represents a sequence of states $\tau = (s_0 \rightarrow s_1 \rightarrow \ldots \rightarrow s_n)$ originating from the initial state $s_0$ and culminating in a sink state $s_n$. $n$ is the trajectory length. Each transition from parent of $s_t$ to $s_{t+1}$ is dictated by the admissible action set $\mathcal{A}(s_t)$ at each time step $t$. The complete set of trajectories, denoted by $\mathcal{T}$, encapsulates all such sequences.
	
	\subsection{Problem Set-up}
	
	The trajectory flow, represented by the function $F:\mathcal{T}\rightarrow \mathbb{R}_{\geq 0}$, delineates the unnormalized probability flux along each complete trajectory from the source to a sink. The flow through state $s$ is the total sum of the flows of all trajectories that include state $s$. Therefore, to estimate the flow passed through a specific state $s$, the flow function for the state can be defined as:
	$$
	F(s)=\sum_{\tau\in\mathcal{T}:s\in\tau}F(\tau).
	$$
	
	Similarly, the flow through edge $s\rightarrow s'$ is the total sum of the flows of all trajectories that include edge $s\rightarrow s'$. To describe the flow for a specific edge $s\rightarrow s'$, the flow is:
	$$
	F(s\rightarrow s')=\sum_{\tau\in\mathcal{T}:(s\rightarrow s')\in\tau}F(\tau).
	$$
	
	In particular, the sum of all the trajectories' flow can be expressed as $Z$. The total flow is the sum of the flows of all trajectories. It is also equal to the flow function of initial state $s_0$.  It is shown in the formula:        
	$$
	Z=\sum_{\tau\in\mathcal{T}}F(\tau)=F(s_0).
	$$
	
	When the flow function has been defined, the probability of a complete trajectory $\tau$ can be defines. The probability of a complete trajectory $\tau$ is is the ratio of the flow passing through trajectory $\tau$ to the total flow. It is expressed as:
	$$
	P(\tau)=\frac{F(\tau)}{Z}.
	$$
	
	To demonstrate the transition dynamics more clearly, the concepts of forward and backward probability are introduced. For an edge $s\rightarrow s'$, $s'$ is the forward state of $s$. Consequently, $s$ is the backward state of $s'$. The forward probability for edge $s\rightarrow s'$ is the ratio of the flow passing through edge $s\rightarrow s'$ to the flow passing through state $s$. It is denoted as:
	$$
	P_F(s'|s)=\frac{F(s\rightarrow s')}{F(s)}.
	$$
	
	Similarly, the backward probability for edge $s\rightarrow s'$ is the ratio of the flow passing through edge $s\rightarrow s'$ to the flow passing through state $s'$. The formula is:
	$$
	P_B(s|s')=\frac{F(s\rightarrow s')}{F(s')}.
	$$
	
	For $s\rightarrow s'$, $s$ is the child of $s'$ and $s$ is the parent of $s'$. Distributions $P_F(\cdot|s)$ are defined over the children of every non-terminal state $s$, alongside a constant $Z$. Then a trajectory flow $F(\tau)$ is considered Markovian if for any complete trajectory $\tau$, the probability of the trajectory follows the product of transition probabilities: 
	$$
	P_F(\tau=( s_0 \rightarrow s_1 \rightarrow \ldots \rightarrow s_n))=\prod_{t=1}^{n}P_F(s_t|s_{t-1}).
	$$
	
	These transition probabilities, denoted as $P_F(s_{t+1}|s_t)$, constitute a forward policy enabling the sampling of complete trajectories. Additionally, we can also consider $P_B(s_{t-1}|s_t)$ as a backward policy.
	
	A key aspect of GFlowNet is the fulfillment of the flow-matching constraint. In particular, an edge flow $F$ is considered as an $R$-edgeflow if it adheres to the reward constraint. The combination of these constraints defines an $R$-flow. $R$-flow is characterized by conditions such as balanced flows into and out of each state. Accordingly, the equivalence between flow directed towards the terminal state $s_f$ and the associated reward is also part of the conditions. The flow-matching constraint for state $s$ means equality of inflows and outflows of $s$. For all the states, the flow-matching constraint can be expressed as:
	$$
	\forall s\in \mathcal{S}, \sum_{s'\in Par(s)}F(s'\rightarrow s)=\sum_{s''\in Child(s)}F(s\rightarrow s'').
	$$
	
	For terminating states, which are the direct parents of terminal states, the outflow is equal to the reward function. The equation is:
	$$
	F(s\rightarrow s_f)=R(s).
	$$
	
	To train a parameterized model of edge flows satisfying the  regular flow-matching constraint and the reward constraint, a loss function $\mathcal{L}$ is formulated. $\delta$ is a hyperparameter, and this notation will also be used in the following loss functions. Several families of losses have been introduced, including the Flow Matching loss (FM) (Bengio et al., 2021a), the Detailed Balance loss (DB) (Bengio et al., 2021b), and the Trajectory Balance loss (TB) (Malkin et al., 2022). These losses ensure the minimization of discrepancies between the learned sampling distribution and the target distribution.
	
	The FM loss is the discrepancy of the inflows and outflows of certain state $s$:
	$$
	L_{FM}(\hat{F},s')=\left\{
	\begin{matrix}
		\left( \log \left( \frac{\delta + \sum_{s\in Par(s')}\hat{F}(s\rightarrow s')}{\delta + R(s') + \sum_{s''\in Child(s')\setminus\{s_f\}}} \right) \right)^2 & \text{if } s' \ne s_f, \\
		0 & \text{otherwise}
	\end{matrix}\right.,
	$$
	FM loss is stage-decomposable, which means that the total FM loss function is decomposed as the sum of all the states' FM loss functions. The equation is: 
	$$
	\mathcal{L}_{FM}(\hat{F})=\sum_{s\in \mathcal{S}}L_{FM}(\hat{F},s).
	$$
	
	The DB loss is paying attention to edges. The DB loss describes the differences between two types of expressions of an edge flow. For an edge $s\rightarrow s'$, the flow can be denoted as $\hat{F}(s)\hat{P}_F(s'|s)$ or $\hat{F}(s')\hat{P}_B(s|s')$. This can be shown in the definitions of forward and backward probabilities. The DB loss can be expressed as:
	$$
	L_{DB}(\hat{F},\hat{P}_F,\hat{P}_B,s')=\left\{
	\begin{matrix}
		\left( \log \left( \frac{\delta + \hat{F}(s)\hat{P}_F(s'|s)}{\delta + \hat{F}(s')\hat{P}_B(s|s')} \right) \right)^2 & \text{if } s' \ne s_f, \\
		\left( \log \left( \frac{\delta + \hat{F}(s)\hat{P}_F(s'|s)}{\delta + R(s)} \right) \right)^2 & \text{otherwise}
	\end{matrix}\right.,
	$$
	DB loss is edge-decomposable, which means that the total DB loss function is decomposed as the sum of all the edges' DB loss functions. The equation is: 
	$$
	\mathcal{L}_{DB}(\hat{F},\hat{P},\hat{P}_B)=\sum_{s\rightarrow s'\in \mathbb{A}}L_{DB}(\hat{F},\hat{P},\hat{P}_B,s\rightarrow s').
	$$
	
	The TB loss focuses on complete trajectories. The TB loss is the discrepancy between two representations of a Markovian trajectory flow. $\hat{Z} \prod_{t=1}^{n+1} \hat{P}_{F}(s_{t}|s_{t-1})$ is the representation using forward probabilities. $R(s_n) \prod_{t=1}^{n} \hat{P}_{B}(s_{t-1}|s_t)$ is the representation using backward probabilities. The TB loss function for trajectory $\tau$ is:
	$$
	\forall \tau=(s_0,...,s_{n+1}=s_f)\in\mathcal{T},
	$$
	$$
	L_{TB}(\hat{Z},\hat{P}_F,\hat{P}_B,\tau)=\left( \log \left( \frac{\hat{Z} \prod_{t=1}^{n+1} \hat{P}_{F}(s_{t}|s_{t-1})}{R(s_n) \prod_{t=1}^{n} \hat{P}_{B}(s_{t-1}|s_t)} \right) \right)^2.
	$$
	
	TB loss is trajectory-decomposable, which means that the total TB loss function is decomposed as the sum of all the complete trajetories' TB loss functions. The equation is: 
	$$
	\mathcal{L}_{TB}(\hat{Z},\hat{P},\hat{P}_B)=\sum_{\tau\in \mathcal{T}}L_{TB}(\hat{Z},\hat{P},\hat{P}_B,\tau).
	$$
	
	The sampling distribution acquired through GFlowNet is labeled as $p(x)$. It is derived by initiating sampling from $s_0$ and continuously selecting $P_{F}(s_{t+1}|s_t)$. Finally, it will  reach a terminal state $x$. The learning objectives aim to align $p(x)$ with the target distribution, denoted as the proportion of state $x$'s reward function to the sum of reward functions:
	$$
	p^{*}(x)\triangleq \frac{R(x)}{\sum_{\mathcal{X}}R(x)}.
	$$
	
	GFlowNet manifests as a learning algorithm governed by parameters $\theta$. The algorithm encompasses a model of a Markovian flow $F_{\theta}$ and an associated objective function. The configuration of the flow model is uniquely specified through various parameters, including the edge flows $F_{\theta}(s\rightarrow s')$, the initial state flow $Z_{\theta}=F_{\theta}(s_0)$, and the terminal state flows $F_{\theta}(x)$.
	
	\section{Model}
	
	Inspired by the analogous properties between GFlowNets and decision trees, this section introduces a novel model. This new model places greater emphasis on network structure characteristics. To incorporate network structural  features effectively, the initial step involves identifying the structure's role in the loss function calculation. The identification is specifically through sub-GFlowNet delineation. Subsequently, defining the loss function for each sub-structure becomes imperative. Finally, integrating these sub-losses necessitates the introduction of substructure entropy as a weighting mechanism. As a result, the comprehensive loss function is formulated. Minimization of this aggregate loss function trains GFlowNet to optimize the selection and evaluation of candidates. Consequently, predefined criteria will be met.
	
	\subsection{Sub-GFlowNet}
	
	As the state-conditional flow network has been proposed by  Bengio et al. (2023), it is shown that subflow network can be taken as the substructure to better learn about the GFlowNet. State-conditional flow network has some assumptions about the terminating flows. To be more specific, a flow network given by a DAG $G=(\mathcal{S},\mathbb{A})$ and a flow function $F$ will have a subgraph of $G$ denoted as $G_s$ for each state $s\in\mathcal{S}$. $G_s$ contains all the states which can be reached from $s$. In other words, the starting state of $G_s$ is not the original initial state $s_0$ but $s$. Bengio et al. designed a conditional flow function $F:\mathcal{S}\times\mathcal{T}\rightarrow R^{+}$, where $\mathcal{T}=\Cup_{s\in\mathcal{S}}\mathcal{T}_s$ and $\mathcal{T}_s$ is the set of complete trajectories in $G_s$. The most important assumption about the flow function is $F_s(s'\rightarrow s_f)=F(s'\rightarrow s_f)$. As a result, the flow of the edges will be changed completely. Thus, the value of forward and backward probability will also be different from those of the original GFlowNet sequentially.  
	
	To simplify the problem and test the effects of the sub-GFlowNet weighting scheme more conveniently, the assumption has been changed. The definitions are provided to better illustrate the new weighting scheme.
	
	\textbf{Definition 1} The set of all the states in the flow network $G=(\mathcal{S},\mathbb{A})$ is $\mathcal{S}$. \textbf{A subflow network state} is defined if the state $s$ has at least 2 outflow edges. As the subgraph of $G$, $G_s$ takes $s$ as the initial state and contains all the complete trajectories in $G_s$. The set of all the complete trajectories in $G_s$ is  denoted as $\mathcal{T}_s$. The subflow function is expressed as $F_s$. For $s$, the outflows are equal in $G$ and $G_s$:
	$$F_s(s\rightarrow s')=F(s\rightarrow s'), \forall s'\in Child(s).$$
	
	The forward probability of the subflow network generated by $s$ is denoted as $P_{F}^{s}$. The forward probabilities of $s$ are also the same in both $G$ and $G_s$:
	$$P_{F}^{s}(s'\rightarrow s'')=P_{F}(s'\rightarrow s''), \forall s',s''\geq s.$$ 
	
	The set of all the states having at least 2 child states is $\mathcal{S}^{*}$.
	
	The definition shows that the scheme only focuses on the intermediate states with branches. In addition, the substructure is taken as an independent structure.
	
	\subsection{Sub-GFlowNet Loss}
	Based on the introduction of various losses, the crucial part about the loss function is the unit where the total loss function can be decomposed. In the new weighting scheme, the loss function is subflow network-decomposable. Essentially, this implies that the loss function is state-decomposable. Next definition is about the loss function of each substructure.
	
	\textbf{Definition 2} For a subflow network generated by $s$, it can be seen as a new glow network having a new initial point $s$. The methods of calculating the loss function of the original flow network can also be applied to the subflow network. The set of subflow network states is defined as $\mathcal{S}_{sub}$ and the corresponding $Z$ is denoted as $Z_s$. For any state $s\in\mathcal{S}_{sub}$, the loss function can be defined as following. First, for every sub-GFlowNet $G_s$, the loss function is defined by TB loss functions. Therefore, the loss function should first be defined over every complete trajectory in $\mathcal{T}_s$. 
	$$\forall\tau=(s_0,...,s_{n+1}=s_f)\in\mathcal{T}_s,$$
	$$L_{SubGFlowTB}(Z_s,P_{F}^{S},P_{B}^{S},\tau)=(log(\frac{Z_s{\textstyle \prod_{t=1}^{n+1}P_{F}^{s}(s_{t}|s_{t-1})}}{R(s_n){\textstyle\prod_{t=1}^{n}P_{B}^{s}(s_{t-1}|s_t)}}))^{2}.$$
	
	Then for every sub-GFlowNet $G_s$, the loss function is the sum of TB loss functions of all the complete trajectories in $\mathcal{T}_s$.
	
	$$L_{SubGFlowTB}(Z_s,P_{F}^{S},P_{B}^{S},s)=\sum_{\tau\in\mathcal{T}_s}L_{SubGFlowTB}(Z_s,P_{F}^{S},P_{B}^{S},\tau),$$
	
	Finally, in this situation the loss function is subGFlowNet-decomposable. It means that the total sub-GFlowNet loss function can be decomposed as the sum of all the sub-GFlowNets'  loss functions:
	
	$$\mathcal{L}_{SubGFlowNet}(Z,P_{F},P_{B})=\sum_{s\in\mathcal{S}_{sub}}L_{SubGFlowTB}(Z_s,P_{F}^{S},P_{B}^{S},s).$$
	As a result, the total loss function can also be seen as state-decomposable.
	
	This new scheme takes the subflow network into consideration, rather than only paying attention to the discrete trajectories. However, this scheme also ignores the weights of each sub-loss function. One of the existed weighting scheme related with trajectory balance has been proposed by Madan et al. (2023) This paper takes total loss function as subGFlowNet-decomposable, so a new weighting scheme should be introduced. To be specific, the entropy of the subflow network will be taken as the weight of each sub-loss function. This strategy is motivated by some concepts of decision tree.
	
	\textbf{Definition 3} In the subflow network $G_s$ generated by $s$, the set of corresponding  terminal states is $\mathcal{S}_{f}^{s}$. The set of states directly connected with states in $\mathcal{S}_{f}^{s}$ is denoted as $\mathcal{S}'$. The \textbf{entropy of the subflow network} is denoted as followed:
	$$Ent(G_s)=-\sum_{s'\in\mathcal{S}',s_f\in\mathcal{S}_{f}^{s}}(\frac{F_s(s'\rightarrow s_f)}{Z_s})log(\frac{F_s(s'\rightarrow s_f)}{Z_s}).$$
	
	Based on the definition of subflow network entropy, the new total loss function under the weighting scheme can be defined as follows:
	$$\mathcal{L}_{SubGFlowNet}(Z,P_{F},P_{B})=\frac{\sum_{s\in\mathcal{S}_{sub}}Ent(G_s)L_{SubGFlowTB}(Z_s,P_{F}^{S},P_{B}^{S},s)}{\sum_{s\in\mathcal{S}_{sub}}Ent(G_s)}.$$
	\section{Experiment}
	The efficacy of the trajectory balance loss function as a primary training objective has been established. Besides, the sub-trajectory loss is recognized as an alternative weighting scheme. As a result, this study systematically assesses the performance of the proposed sub-GFlowNet loss function. Specifically, this research compares it against the trajectory balance and sub-trajectory loss functions. The evaluation encompasses experiments conducted across diverse scenarios. The experiments include hypergrid environments of varying dimensions and sizes, as well as the molecule synthesis task.
	
	\subsection{Numerical Stimulation}
	\textbf{Hypergrid environment}
	
	In this section, this paper delves into a synthetic hypergrid environment introduced in Bengio et al.(2021). While this task is less complex compared to others under examination, its inclusion is necessary for comprehensiveness. In addition, it is able to elucidate various noteworthy behaviors.
	
	In this constructed environment, the nonterminal states $\mathcal{S}^{\circ}$ form a hypergrid of dimensionality $D$, with each side having a length of $H$: $$\mathcal{S}^{\circ}=\{(s^1,...,s^D)|s^d\in\{0,1,...,H-1\},d=1,...,D)\},$$
	where actions involve incrementing one coordinate within a state by 1. The coordinates will not exceed the grid boundaries. The initial state is set to (0,...,0). Additionally, for each nonterminal state $s$, there exists a termination action that transitions to a corresponding terminal state $s^T$. The reward at a terminal state $s^T=(s^1,...,s^d)^T$ is given by:
	$$R(s^T)=R^0+0.5\prod_{d=1}^{D}\mathbb{I}[|\frac{s^d}{H-1}-0.5|\in(0.25,0.5)]+2\prod_{d=1}^{D}\mathbb{I}[|\frac{s^d}{H-1}-0.5|\in(0.3,0.4)].$$
	where $\mathbb{I}$ denotes the indicator function and $R_0$ represents a constant parameter influencing exploration difficulty. This reward function exhibits peaks of magnitude $2.5+R_0$ near the corners of the hypergrid. The peaks are  surrounded by plateaux of height $0.5+R_0$. These plateaux are separated by wide troughs with a reward of $R_0$. The objective of this environment is to assess the capacity of a GFlowNet to generalize from visited states. Additionally, the objective also includes inferring the existence of yet-unvisited modes.
	
	This study investigates grid environments of dimensions $2$, $3$, and $4$, with grid lengths set at $H=8$, $16$, and $32$ respectively. A uniform backward policy is adopted. The policy is consistent with prior research methodologies. In subsequent visualizations, the trajectory balance training objective is depicted in red, while the sub-trajectory loss is represented in green. The novel weighted sub-GFlowNet loss, introduced in this paper, is illustrated in blue.
	
	\begin{figure}[htbp]
		\centering
		\begin{subfigure}[b]{0.32\textwidth}
			\centering
			\includegraphics[width=\textwidth]{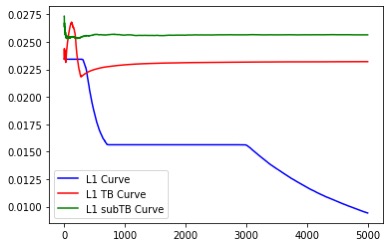}
			\caption{dim=2,horizon=8}
			\label{fig:subfig1.1}
		\end{subfigure}
		\hfill
		\begin{subfigure}[b]{0.32\textwidth}
			\centering
			\includegraphics[width=\textwidth]{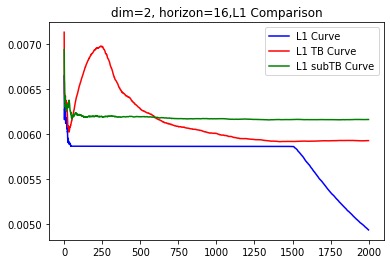}
			\caption{dim=2,horizon=16}
			\label{fig:subfig1.2}
		\end{subfigure}
		\hfill
		\begin{subfigure}[b]{0.32\textwidth}
			\centering
			\includegraphics[width=\textwidth]{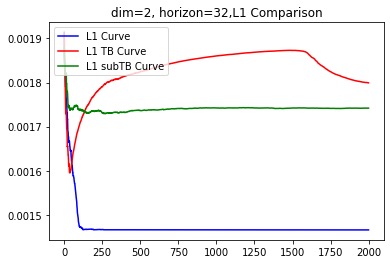}
			\caption{dim=2,horizon=32}
			\label{fig:subfig1.3}
		\end{subfigure}
		\begin{CJK*}{UTF8}{gbsn}
			\caption{Comparison of L1 Distance Between Empirical and True Distributions as Horizon Varies for GFlowNet with Different Loss Functions in dim=2
			}
		\end{CJK*}
		\label{fig:1}
	\end{figure}
	
	As depicted in Figure 1, subfigure (a) demonstrates that the GFlowNet with the novel training objective achieves the best performance in the $8\times 8$ hypergrid. The novel GFlowNet  exhibits the highest convergence speed and the lowest L1 distance, while the subTB GFlowNet performs the worst. Additionally, the trajectory balance GFlowNet's curve appears ragged. Subfigure (b) illustrates similar results to those in subfigure (a), and subfigure (c) shows that the plot line of the newly proposed GFlowNet rapidly stabilizes and levels off at a plateau. In the $32\times 32$ hypergrid, the TB GFlowNet performs the worst in terms of both stability and L1 distance value. Collectively, these subfigures highlight the differences in performance between the various GFlowNets in 2-dimensional grids with different horizons. Analysis of the 2-dimensional grid experiments reveals that the proposed weighting scheme yields superior performance. Specifically, the performances are characterized by accelerated convergence rates and reduced terminal L1 distance between empirical and real distributions. Furthermore, the proposed objective exhibits greater stability compared to alternative objectives.
	
	\begin{figure}[htbp]
		\centering
		\begin{subfigure}[b]{0.32\textwidth}
			\centering
			\includegraphics[width=\textwidth]{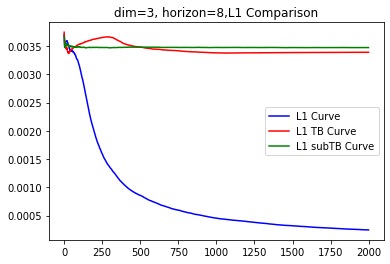}
			\caption{dim=3,horizon=8}
			\label{fig:2.1}
		\end{subfigure}
		\hfill
		\begin{subfigure}[b]{0.32\textwidth}
			\centering
			\includegraphics[width=\textwidth]{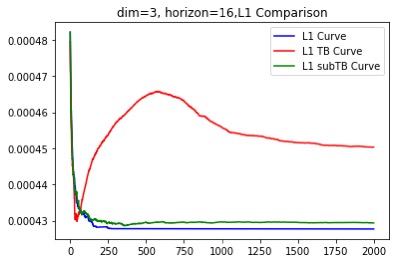}
			\caption{dim=3,horizon=16}
			\label{fig:2.2}
		\end{subfigure}
		\hfill
		\begin{subfigure}[b]{0.32\textwidth}
			\centering
			\includegraphics[width=\textwidth]{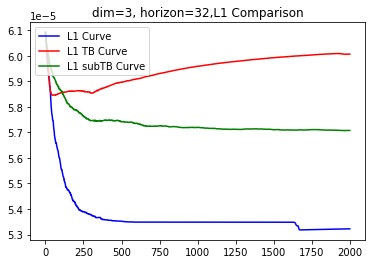}
			\caption{dim=3,horizon=32}
			\label{fig:2.3}
		\end{subfigure}
		\begin{CJK*}{UTF8}{gbsn}
			\caption{Comparison of L1 Distance Between Empirical and True Distributions as Horizon Varies for GFlowNet with Different Loss Functions in dim=3
			}
		\end{CJK*}
		\label{fig:2}
	\end{figure}
	
	In Figure 2, as depicted in subfigure (a) of the $8\times 8\times 8$ hypergrid, both the TB and subTB GFlowNets rapidly stabilize. However, both of their plateau L1 values are significantly higher than that of the newly proposed GFlowNet. Similarly, the TB GFlowNet exhibits a more unstable curve. In subfigure (b), the performances of the two GFlowNets with different weighting schemes are comparable, although the TB GFlowNet initially performs best but quickly rebounds. Consequently, TB GFlowNet results in the worst performance in the $16\times 16\times 16$ grid. Lastly, in subfigure (c), the novel GFlowNet clearly outperforms the other two methods. The TB GFlowNet remains the least effective. These subfigures collectively underscore the performance differences among various GFlowNets in 3-dimensional grids with different horizons. Upon extending the evaluation to the 3-dimensional grid, a decrease in convergence speed relative to the 2-dimensional space is observed. The decrease of speed may be  attributable to the heightened complexity and increased potential trajectories of the 3-dimensional grid. Notwithstanding, newly proposed scheme continues to outperform competing objectives in this environment.
	
	\begin{figure}[htbp]
		\centering
		\begin{subfigure}[b]{0.32\textwidth}
			\centering
			\includegraphics[width=\textwidth]{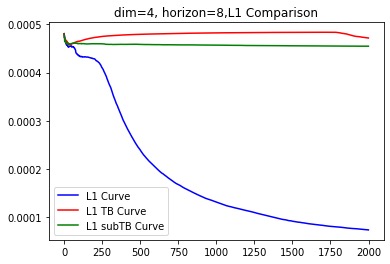}
			\caption{dim=4,horizon=8}
			\label{fig:subfig1}
		\end{subfigure}
		\hfill
		\begin{subfigure}[b]{0.32\textwidth}
			\centering
			\includegraphics[width=\textwidth]{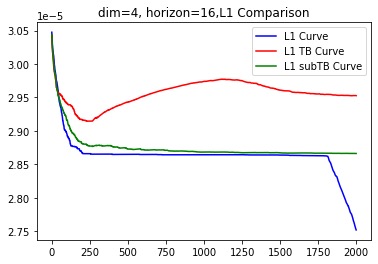}
			\caption{dim=4,horizon=16}
			\label{fig:subfig2}
		\end{subfigure}
		\hfill
		\begin{subfigure}[b]{0.32\textwidth}
			\centering
			\includegraphics[width=\textwidth]{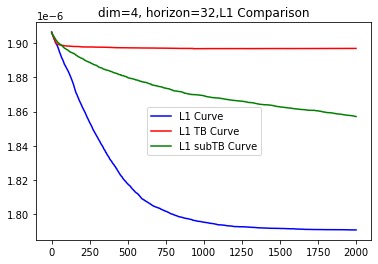}
			\caption{dim=4,horizon=32}
			\label{fig:subfig3}
		\end{subfigure}
		\begin{CJK*}{UTF8}{gbsn}
			\caption{Comparison of L1 Distance Between Empirical and True Distributions as Horizon Varies for GFlowNet with Different Loss Functions in dim=4
			}
		\end{CJK*}
		\label{fig:subfigures}
	\end{figure}
	
	In Figure 3, as shown in subfigure (a), the convergence rate of the GFlowNet employing the sub-GFlowNet loss is slightly slower compared to the same GFlowNet in the $8\times 8\times 8$ hypergrid. However, the new GFlowNet still demonstrates superior performance relative to the other two GFlowNets with different training objectives. The other two curves exhibit comparable performance. In subfigure (b), both subTB and the new GFlowNets perform similarly in the initial 1750 iterations. However, the GFlowNet using the new training loss function continues to decline, achieving the lowest L1 distance. The TB GFlowNet consistently exhibits the poorest performance in stability and L1 distance values. In subfigure (c), the pairwise distinctions among the three are clearly evident. The  GFlowNet utilizing the new training objective achieves the best performance while the TB GFlowNet performs the worst. Similar to Figures 1 and 2, the performance of the newly proposed GFlowNet stands out among the three GFlowNets with different training objectives in 4-dimensional grids with varying horizons. Finally, examination of the 4-dimensional grid corroborates earlier findings regarding convergence speed. The sub-GFlowNet loss emerging as the optimal training objective across dimensions.
	
	\subsection{Real Data}
	\textbf{Molecule synthesis}
	
	In this study, the exploration ventures into the domain of molecule generation, a subject first introduced for GFlowNets in Bengio et al.(2021). The present investigation enriches the existing codebase from Bengio et al.(2021) by incorporating implementations for the sub-GFlowNet loss function. The objective is to generate molecular structures represented as graphs. Meanwhile, the binding affinity to the 4JNC inhibitor of the sEH (soluble epoxide hydrolase) protein should also be minimized. These generated graphs materialize as junction trees assembled from a predefined lexicon of molecular building blocks. The maximum trajectory length is set at 8. The number of actions fluctuating between approximately 100 and 2000. The fluctuation is contingent upon molecular complexity and potential modifications. Consequently, the cardinality of the action space, denoted as $|\mathcal{X}|$, approaches $10^{16}$.
	
	The reward metric is formulated as the normalized negative binding affinity. The affinity is predicted by a surrogate model trained to estimate energies derived from docking simulations. In this research, the Tanimoto index, also known as the Jaccard index, serves as the diversity metric. This index quantifies the degree of overlap between two sets by calculating the ratio of their intersection to their union. Specifically, given two sets $A$ and $B$, the Tanimoto index is computed as follows:
	$$Tanimoto(A,B)=\frac{|A\cap B|}{|A\cup B|}.$$
	Here, $|A\cap B|$ denotes the number of elements in the intersection of sets $A$ and $B$, and $|A\cup B|$ represents the number of elements in the union of sets $A$ and $B$. The Tanimoto index ranges from 0 to 1. A value close to 1 indicates a high degree of overlap and hence high similarity between the two sets. On the contrary, a value close to 0 indicates a low degree of overlap and low similarity. In the realm of chemistry, the Tanimoto index finds common use in measuring the similarity of compounds, particularly in compound screening and drug discovery.
	
	In the following graphs, the green line depicts the evolution trend of the proposed GFlowNet in this paper. The blue line represents the performance of the GFlowNet with sub-trajectory balance as the training objective. Last, the red line represents the conventional TB GFlowNet. In this section, the $\lambda$ for the subtb GFlowNet is set to 0.99. Each training epoch simultaneously generates eight trajectories.
	
	\begin{figure}[H]
		\centering
		\begin{subfigure}[b]{0.49\textwidth}
			\centering
			\includegraphics[width=\textwidth]{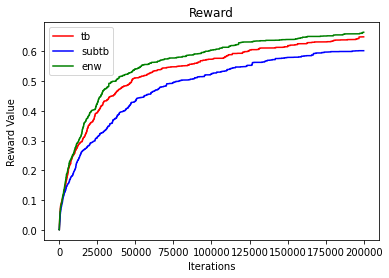}
			\caption{Reward Evolution of GFlowNet under Three Different Training Objectives over 200,000 Iterations}
			\label{fig:subfig1}
		\end{subfigure}
		\hfill
		\begin{subfigure}[b]{0.49\textwidth}
			\centering
			\includegraphics[width=\textwidth]{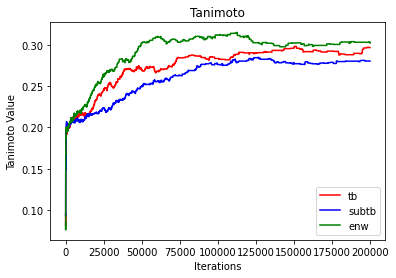}
			\caption{Tanimoto Index Evolution of GFlowNet under Three Different Training Objectives over 200,000 Iterations}
			\label{fig:subfig2}
		\end{subfigure}
		\caption{Comparative Evolution of Performance Metrics in GFlowNet under Three Different Training Objectives over 200,000 Iterations}
		\label{fig:subfigures}
	\end{figure}
	
	It is evident from the results that after 200,000 iterations, the GFlowNet with the new training objective achieves the highest reward value with the fastest converging speed. However, it is also observed that the TB GFlowNet and the entropy-weighted GFlowNet exhibit similar performances regarding reward. In terms of the Tanimoto index, the subtb GFlowNet outperforms the others by generating a more diverse set of molecules.
	
	\section{Discussion}
	This paper introduces a novel GFlowNet with a refined training objective. The proposed GFlowNet emphasizes the significance of weights and network structure over conventional approaches. Specifically speaking, the novel method directs attention towards key elements such as splitting points and sub-GFlowNets generated by bifurcation points. Employing the entropy of sub-GFlowNets as a new weighting scheme, it formulates a corresponding new loss function.
	
	In the hypergrid experiment, the entropy-weighted GFlowNet outperforms both the TB and subTB GFlowNets across various experimental settings. Similarly, in the molecule synthesis task, the entropy-weighted GFlowNet demonstrates success in generating molecules with high rewards, albeit with slightly lower diversity compared to the subTB GFlowNet. It is noteworthy that in a 2-dimensional setting, the subTB GFlowNet consistently performs the worst. However, as the dimensionality increases, the TB GFlowNet exhibits the poorest performance across different horizons. This disparity might be attributed to the expansion of the action space. As the action space expands, the subTB GFlowNet appears to better capture the information of the DAG during the training process. Therefore, a higher efficiency will be achieved bt subTB GFlowNet. Conversely, the results are markedly different when it comes to real data analysis. The action space for molecule synthesis being several orders of magnitude larger than that of the hypergrid experiment. Nevertheless, the TB GFlowNet significantly outperforms the subTB GFlowNet with respect to the predefined reward value. This discrepancy can probably be explained by the imbalance between action space and trajectory length. In detail, the action space in real data analysis is much larger than in numerical simulation. However, the length of each trajectory is limited to 8 blocks, whereas the maximum trajectory length in the simplest grid environment is 16. Thus, the differing results between subTB and TB GFlowNet might be due to the subTB GFlowNet's focus on sub-trajectories. The emphasis better exploits the features of real DAGs when the trajectories are long. Additionally, the molecule synthesis task encompasses two objectives. The two objectives include molecular diversity and predefined rewards. As a result, the evaluation of molecular diversity may also have an unknown impact on the results about rewards.
	
	Despite the promising results, several limitations persist. In hypergrid environments, the GFlowNet with the entropy-weighted loss function clearly outperforms the other two GFlowNets. In the real data analysis, the newly proposed GFlowNet still achieves the best performance in terms of reward. However, the difference between the entropy-weighted GFlowNet and the other two GFlowNets is not pronounced. The subtle differences are  possibly due to the extensive action space. This suggests a limitation: when the action space is vast, the entropy-weighted GFlowNet may not adequately capture the features of the DAG structure. This limitation necessitates further research. Additionally, the molecule synthesis task involves multiple objectives. The results indicate that the newly proposed GFlowNet performs the worst in terms of molecule diversity. This may be attributed to the conflicting requirements of enhancing molecule diversity and improving reward. Nevertheless, addressing multiple objectives in real data tasks remains an area that requires further investigation.
	
	In conclusion, this study underscores the efficacy of the entropy-weighted GFlowNet in diverse experimental setups and tasks. However, addressing the aforementioned limitations and exploring avenues for further refinement remain imperative for advancing the field of flow-based generative models. Future research could explore avenues such as conditional sub-GFlowNets, scalability of training trajectories, and optimization of loss function components. These future studies may propel the efficacy and applicability of GFlowNets in various domains.

\end{document}